# Viability of machine learning to reduce workload in systematic review screenings in the health sciences: a working paper

Muhammad Maaz
Faculty of Health Sciences, McMaster University
Email: maazm@mcmaster.ca

ABSTRACT. Systematic reviews, which summarize and synthesize all the current research in a specific topic, are a crucial component to academia. They are especially important in the biomedical and health sciences, where they synthesize the state of evidence and conclude the best course of action for various diseases, pathologies, and treatments. Due to the immense amount of literature that exists, as well as the output rate of research, reviewing abstracts can be a laborious process. Automation may be able to significantly reduce this workload. Of course, such classifications are not easily automated due to the peculiar nature of written language. Machine learning may be able to help. This paper explored the viability and effectiveness of using machine learning modelling to classify abstracts according to specific exclusion/inclusion criteria, as would be done in the first stage of a systematic review. The specific task was performing the classification of deciding whether an abstract is a randomized control trial (RCT) or not – a very common classification made in systematic reviews in the healthcare field. Random training/testing splits of an n=2042 dataset of labelled abstracts was repeatedly created (1000 times in total), with a model trained and tested on each of these instances. Performance statistics were then averaged across. A Bayes classifier as well as an SVM classifier were used, and compared to non-machine learning, simplistic approaches to textual classification. An SVM classifier was seen to be highly effective, yielding a 90% accuracy, as well as an F1 score of 0.84, and yielded a potential workload reduction of 70%. This shows that machine learning has the potential to significantly revolutionize the abstract screening process.

1. INTRODUCTION

The rate of publication of new research in healthcare, and indeed perhaps other fields, is such that it is difficult for researchers and practitioners to keep up to date (1, 2). Because of this, the importance of systematic reviews cannot be overstated. However, even here, the incredible rate of academic output becomes a problem as reviewers spend many long and tedious hours sifting through thousands and thousands of papers. Often, reviewers need to perform simple evaluations which, despite their simplicity, can become laborious when compounded by the sheer number of papers being reviewed. One of these simple tasks, encountered in many abstract screenings, is deciding whether a paper is a randomized control trial (RCT) or not from its abstract. For such a simple task, the potential for automation to cut down on reviewer's workload is immense.

Systematic reviews pose a unique problem for automation: the nuances and peculiarities of written language make it impossible to make a naive algorithm for classification. Considering the simple problem of classifying an abstract as an RCT, simply searching for the occurrence of the words "RCT" or "randomized control trial" or variants thereof is insufficient. This is because abstracts with those keywords are not always necessarily RCTs (they may simply be outlining past RCTs as a narrative review, or speaking about RCTs in a general sense), nor do all RCTs, unfortunately, make it clear that they are randomized control trials in the abstract by using those keywords. Hence, we must start looking at features like context of the usage of those keywords, or other indirect ways that the abstract can suggest a RCT. Therefore, it is easy to see that the complexity of the problem quickly exceeds the capacity of such a basic algorithm.

These issues can be largely handled using machine learning methods. This entails the computer being fed some training data, from which it constructs its own algorithm (3). From this algorithm, the machine can make predictions on other abstracts that it has never seen. Machine learning is a very powerful tool which has been used widely for myriad predictive applications, including for example classifying email as spam or not (3). A literature search has revealed that applications of such methods to the field of systematic reviews remains little explored.

## 2. METHODS

*2.1. Overview of procedure*
A dataset of abstracts (n = 2042) which were collected and labelled by a human reviewer (0 representing not an RCT, and 1 representing an RCT) for another ongoing systematic review was used. A random sample of training data comprising 80% of the full dataset (n = 1633) was taken, and the remaining 20% (n = 409) was reserved for testing. The training data was processed before the machine was trained on it (the method of which will be described below). The machine was then tested on the testing set, and its predictions compared to the human reviewer's labelling of that data. Note that during training, the machine sees the processed abstract as well as the human reviewer's classification, but during testing, the machine only sees the processed abstract and produces its own classifications, which are then compared to the human reviewer. This process of performing random training/testing splits on the original dataset and then training and testing was performed 1000 times. Statistics regarding the performance of the algorithm were compiled and then averaged across the run-throughs.

*Figure 1. Splitting of dataset with mean representation of labels in training and testing sets. 0 represents non-RCT and 1 represents RCT.*

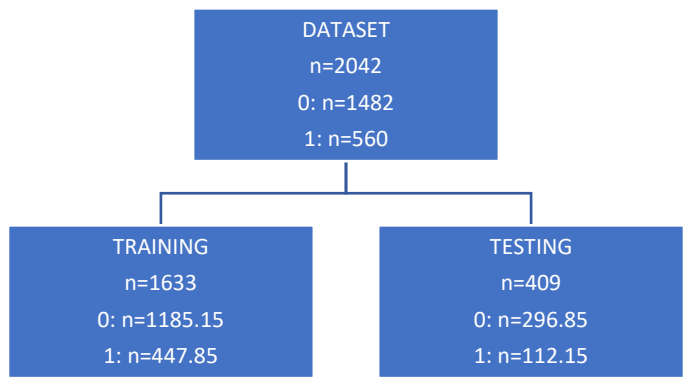

## 2.2. Processing of abstracts

As computers are not particularly good at dealing with text, some sort of processing needs to be performed to make the dataset more palatable (4). The bag-of-words model was used, which involves converting a piece of text into a matrix of its unique terms and their corresponding frequency (4). After this, stop words, which are ubiquitous terms like "the" and "a", were removed, to not burden the machine with words that are largely irrelevant to its understanding (4). The last step was transforming the bag-of-words matrix through a TFIDF transformation. TFIDF (term frequency inverse document frequency) is a measure that scales upwards with the frequency with which a word appears in a document but is offset by its frequency in the entire corpus, to account for the fact that certain words appear more frequently in general (5). The final matrix after all this processing was used as the feature set to train the machine.

## 2.3. Algorithms used

Two different machine learning algorithms were used: a Bayes classifier and a SVM (support vector machine) in Python using the ScikitLearn package (6). Each of these classifiers use distinct mathematical optimizations in the construction of their algorithm which are out of the scope of this paper. As a sort of negative control, two non-machine learning algorithms were also constructed for comparison purposes. One, which we have called the "nuclear" algorithm, simply labels everything as 0 (non-RCT). The other, which we have called the "basic" algorithm, merely looks for the occurrence of the words "RCT" or "randomized" in the given abstract and will label the abstract as an RCT if such an occurrence is found.

## 3. RESULTS

We summarize the results of our algorithm below, and expand on the implications in the discussion.

Table 1. Confusion matrix of "nuclear" algorithm in format mean(std dev).

|  |  | PREDICTED | |
|---|---|---|---|
|  |  | NOT | RCT |
| ACTUAL | NOT | 296.85(8.07) | 0.00(0.00) |
| ACTUAL | RCT | 112.15(8.07) | 0.00(0.00) |

Table 2. Confusion matrix of "basic" algorithm in format mean(std dev).

|  |  | PREDICTED | |
|---|---|---|---|
|  |  | NOT | RCT |
| ACTUAL | NOT | 257.06(8.83) | 39.79(5.26) |
| ACTUAL | RCT | 51.78(6.33) | 60.37(6.32) |

Table 3. Confusion matrix of Bayes machine learning algorithm in format mean(std dev).

|  |  | PREDICTED | |
|---|---|---|---|
|  |  | NOT | RCT |
| ACTUAL | NOT | 296.07(8.00) | 0.78(0.84) |
| ACTUAL | RCT | 88.25(10.19) | 23.90(4.09) |

Table 4. Confusion matrix of SVM machine learning algorithm in format mean(std dev).

|  |  | PREDICTED | |
|---|---|---|---|
|  |  | NOT | RCT |
| ACTUAL | NOT | 272.80(8.71) | 24.05(4.66) |
| ACTUAL | RCT | 14.09(3.73) | 98.06(7.83) |

Table 5. Summary performance statistics of different algorithms. Accuracy and F1 score range from 0 to 1.

| Algorithm | Accuracy (mean(std dev)) | F1 score (mean(std dev)) |
|---|---|---|
| Nuclear | 0.7258(0.0197) | N/A |
| Basic | 0.7761(0.0190) | 0.5678(0.0365) |
| Bayes | 0.7823(0.0247) | 0.3498(0.0590) |
| SVM | 0.9068(0.0132) | 0.8367(0.0238) |

Figure 2. Summary performance statistics of different algorithms. Error bars show standard deviation.

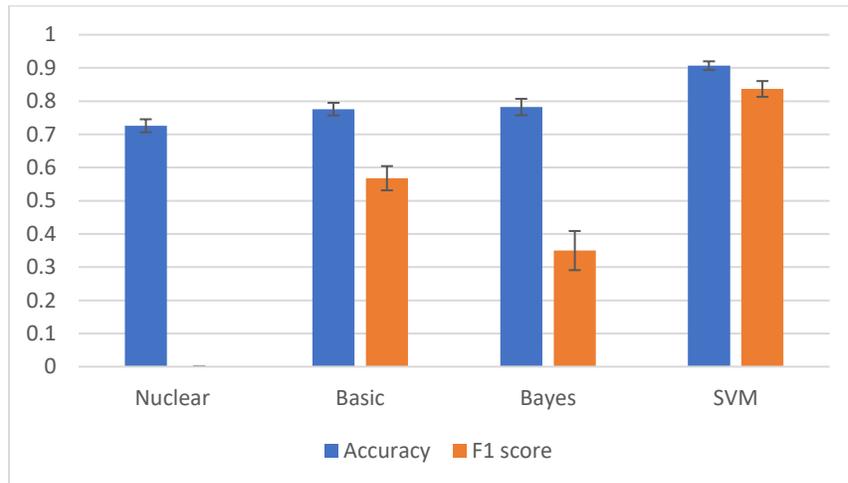

Table 6. Unpaired t-test of F1 scores between different algorithms. No comparisons can be made with nuclear due to undefined F1 score.

|  | Nuclear | Basic | Bayes | SVM |
|---|---|---|---|---|
| Nuclear | - | - | - | - |
| Basic | - | - | p < 0.0001 | p < 0.0001 |
| Bayes | - | p < 0.0001 | - | p < 0.0001 |
| SVM | - | p < 0.0001 | p < 0.0001 | - |

## 4. DISCUSSION

Initially, only a Bayes classifier was used as the machine learning algorithm. However, similar performance was observed compared to the nuclear and basic algorithms. This would ostensibly seem as if machine learning is of no use, as such naïve algorithms can achieve such similar results, with far more simplicity of implementation. It is important to note that despite statistically significant differences in the F1 scores of basic and Bayes, this is misleading as practically speaking both scores are subpar. Similarly, the accuracies of nuclear, basic, and Bayes are all practically similar and subpar.

From this, the question arose as to why the Bayes classifier was performing so poorly. Of course, it is interesting to note that we would not think that the Bayes classifier was performing poorly if not for the simplistic algorithms performing so similarly. The issue with the dataset was clearly the imbalance – that is to say, an imbalanced underrepresentation of RCTs. Looking at Figure 1, in the full dataset, only 27% of the abstracts are RCTs, and, in fact, this proportion was conserved in the randomized training and testing datasets. This is why the nuclear algorithm which simply labels every abstract it encounters as 0 is able to achieve such high accuracy. However, it is obvious such an algorithm is useless as it delivers zero true positives and hence zero recall and zero precision (hence the undefined F1 score, which is the harmonic mean of recall and precision (6)).

Therefore, there needs to be a way to solve for this class imbalance. The SVM classifier offers a solution, as it can assign a rebalancing weight such that data is weighed inversely to the frequency of the class to which it belongs (6,7). Hence, an RCT is considered more important to the classifier during the training phase precisely because it appears less frequently. The hope is that this will balance out the class imbalance and deliver a better performing classifier. Therefore, a SVM classifier was constructed, and indeed this hope was realized.

The SVM classifier yielded very strong results in all regards. It achieved approximately a 91% accuracy – although it is important to note that this is a crude measure due to the imbalance in the dataset, as evidenced by the 73% accuracy of the nuclear algorithm. It is better then to look at the confusion matrices of each algorithm. The SVM not only yielded many more true positives, but also fewer false negatives. It is arguable that in the abstract screening stage, reducing false negatives is priority as one does not want to over-exclude in this stage. If any non-RCTs manage to make through, they can still be excluded during the full text screening; hence, false positives are not a huge concern. Interestingly enough, while SVM had fewer false positives than the basic algorithm, it had many more false positives than the Bayes classifier, but this is perhaps more so due to the tendency of the Bayes classifier to label abstracts as non-RCT, as shown by its closeness to the nuclear algorithm. As a composite score of the confusion matrices, F1 score was calculated, which is a more robust measure of performance than accuracy. The SVM classifier exhibited a practically significant higher F1 score than all other algorithms used, reflecting its much higher precision and recall. From these statistics, it is clear the SVM classifier outperformed all other algorithms in the relevant statistics.

Of course, the original objective of this study was to examine the viability of using a machine learning model to classify abstracts. In this regard, looking at the SVM classifier, the potential for workload reduction is huge. Looking at Table 4, classifying 70% of the testing set as non-RCTs, and a false omission rate (i.e. the proportion of rejects that were falsely rejected) of only 5%, we see that simply relying on the machine learning classifier to exclude abstracts would yield a massive workload reduction, leaving only 30% of abstracts for human reviewers to review for further inclusion/exclusion, or to pass on to full text screening. Given the thousands of abstracts typically reviewed in systematic reviews, the time saved from such a reduction is immense.

It is remarkable that such a simple characteristic can discriminate abstracts so well. In the future, we can begin to construct multi-class classification models which can label abstracts or full texts on many various characteristics. It is our hope that the abstract screening process can be automated fully or at least to a greater extent. Such automated techniques would help reviewers and researchers better keep up with today's massive academic output. A machine learning model trained on a large corpus can be exported for anyone to use on their own set. Thus, it is not a stretch to envision a website or a software package with such preloaded models in which researchers could upload their own abstract sets for classification. It is interesting to note, however, that a software program that automates the systematic review process to deal with the exponential growth in academic output would paradoxically increase academic output as it would make systematic reviews easier to complete. Regardless, this paper has clearly demonstrated that a machine learning model can largely automate most of the tedious labour typically handled by human reviewers.

## 5. CONCLUSION

Systematic reviews are laborious but necessary. The potential for automation to reduce workload, especially for simple tasks, is massive. A Bayes classifier and a SVM classifier were constructed and tested. The SVM classifier, which corrected for the class imbalance inherent in the dataset, exhibited stellar performance on the testing set, demonstrating the potential for a workload of 70%, merely based on such a simple characteristic as whether the abstract is an RCT or not. This is an exciting first step in the application of machine learning methods to the systematic review process, which, if further developed, could revolutionize the systematic review process by allowing researchers to be able to better handle the massive amount of papers that they have to read during a systematic review.

## 6. ACKNOWLEDGEMNTS

The author would like to acknowledge Dr. Ahmed Negm for providing the abstract dataset used in this study.